\documentclass[journal]{journal}
\usepackage{multicol}
\usepackage{float}
\usepackage{graphicx}
\usepackage[utf8]{inputenc}
\usepackage{color}
\usepackage{url}
\usepackage{amsmath}
\usepackage[normalem]{ulem}
\newcommand{\nothing}[1]   {}

\begin{document}
%
\title{Single-Camera Basketball Tracker through Pose and Semantic Feature Fusion}
%
%
%

\author{Adrià Arbués-Sangüesa$^{1,2}$, Coloma Ballester$^1$, Gloria Haro$^1$ \\ $^1$ Universitat Pompeu Fabra (Barcelona, Spain) and $^2$ FC Barcelona}

\maketitle
\thispagestyle{empty}

\begin{abstract}
Tracking sports players is a widely challenging scenario, specially in single-feed videos recorded in tight courts, where cluttering and occlusions cannot be avoided. This paper presents an analysis of several geometric and semantic visual features to detect and track basketball players. An ablation study is carried out and then used to remark that a robust tracker can be built with Deep Learning features, without the need of extracting contextual ones, such as proximity or color similarity, nor applying camera stabilization techniques. The presented tracker consists of: (1) a detection step, which uses a pretrained deep learning model to estimate the players pose, followed by (2) a tracking step, which leverages pose and semantic information from the output of a convolutional layer in a VGG network. Its performance is analyzed in terms of MOTA over a basketball dataset with more than 10k instances. 
\end{abstract}


\IEEEpeerreviewmaketitle

\section{Introduction} \label{sec:Intro}

Basketball professional European courts are 28 meters large and 14 meters wide, and 10 players (plus 3 referees) interact on it following complex patterns that help them accomplishing their goal, which can be either scoring or preventing the other team to score. These tactical plays include several kinds of movement, which might involve players moving together and close to each other, thus generating space and advantages. Given this scenario, and being aware that there is not an established multi-camera array set in these courts for tracking purposes (because of its cost and the low height of stadiums), broadcasting cameras are the main source of basketball video content. Usually, these cameras are set in the middle of the court in the horizontal axis, and camera operators just perform some panning or zooming during the game right at the same spot. 

{Given this video feed, a  tracking-by-detection algorithm is adopted: first, potential players are detected, and then, features are extracted and compared, quantifying how much do players in different frames resemble. Then, players are tracked by obtaining a matrix with all similarities and minimizing the total cost of assignments. Several kind of features for establishing the similarities are evaluated:}

\nothing{
\textit{
Given this video feed, several choices can be made in order to build a multi-tracker: 
\begin{itemize}
    \item A simultaneous tracking-by-detection algorithm can be designed\textcolor{blue}{/is adopted/is proposed (amb can no queda clar si ho fem o no)}: first, potential players have to be detected, and then, features are extracted and compared, quantifying how much do players in different frames resemble. By obtaining a matrix with all similarities and minimizing the total cost of assignments, players can be tracked. 
    \item An optical flow method might be used instead, by computing displacement vectors for every pixel/object. In this way, and combining with a (possibly spatio-temporal) segmentation method, players can be localized and their movements estimated. \textcolor{blue}{aquest punt em resulta confus perquè no fem res d'això i de fet no sé si algun paper de tracking mitjanament competitiu ho fa} 
\end{itemize}
In this article, a tracker based on detections is presented, and several kind of features are tested:}
}
\begin{itemize}
    \item Geometrical features, which might involve relative distances (in screen-coordinates and expressed in pixels) between detected objects. 
    \item Visual features, which may quantify how different boxes look alike by comparing RGB similarity metrics in different small neighborhood patches. 
    \item Deep learning features, which can be obtained by post-processing the output of a convolutional layer in a Deep Neural Network.  
\end{itemize}
\nothing{
\textit{Besides, the combination with classical Computer Vision techniques might help improving the trackers' overall performance. For instance, homography estimation can compute transformations within consecutive frames, thus leading to stabilized camera sequences, where distances among frames are considerably reduced.}
}

{Besides, we show that the combination with classical Computer Vision techniques helps improving the trackers' overall performance. In particular, camera stabilization based on homography estimation leads to camera motion compensated sequences where distances of corresponding players in consecutive frames are considerably reduced.}

The aim of this paper is to prove that deep learning features can be extracted and compared with ease, obtaining better results than with classical features. For this reason, an ablation study for different tests in a given scenario is included. The remaining article is distributed as follows: in Section \ref{sec:SoA} related works are described. Later on, in Section \ref{sec:Meth}, the presented methods are detailed, involving the main modules of player and pose detection, feature extraction and matching; moreover, camera stabilization techniques and pose models are considered. Results are shown and discussed in Section \ref{sec:Res}, and conclusions are extracted in final Section \ref{sec:Conc}.

\section{Related Work} \label{sec:SoA}
Multi-object tracking in video has been and still is a very active research area in computer vision. One of the most used tracking strategies is the so-called tracking by detection, which involves a previous or simultaneous detection step~\cite{milan2015joint,ramanathan2016detecting,henschel2018fusion,girdhar2018detect,doering2018joint}.
Some of these works use a CNN-based detector with a tracking step \cite{ramanathan2016detecting,girdhar2018detect}, while others are based on global optimization methods. Among them, a joint segmentation and tracking of multiple targets is proposed in \cite{milan2015joint}, while in 
 \cite{henschel2018fusion}  a full-body detector and a head detector are combined to boost the performance. The authors in 
 \cite{doering2018joint} combine Convolutional Neural Networks (CNNs) and a Temporal-Flow-Fields-based method. 
 Another family of tracking methods which achieve a good compromise between accuracy and speed are based on Discriminant Correlation Filters (DCF). They are based on a first stage where features are extracted and then correlation filters are used. Initially, hand-crafted features like HoG were used and later on different proposals use deep learning features extracted with pretrained networks (e.g. \cite{qi2016hedged}). The results are improved when learning the feature extraction network in an end-to-end fashion for tracking purposes \cite{wang2017dcfnet}. The latest trend is to train deep learning based tracking methods in an unsupervised manner \cite{wang2019learning,wang2019unsupervised}.

On the other hand, pose tracking refers in the literature to the task of estimating anatomical human keypoints and assigning unique labels for each  keypoint across the frames of a video \cite{iqbal2017posetrack,insafutdinov2017arttrack}. 

This paper addresses the problem of tracking basketball players in broadcast videos. This is a challenging scenario where multiple occlusions are present, the resolution of the players is small and there is a high similarity between the different instances to track, specially within the same team members. For a deeper review of players detection and tracking in sports the interested reader is referred to the recent survey \cite{thomas2017computer}. 
The authors of \cite{senocak2018part} also consider basketball scenarios seen from a broadcast camera and they deal with player identification. For that, they propose to use CNN features extracted at multiple scales and encoded in a Fisher vector. 

\section{Proposed Method and Assessment} \label{sec:Meth}
\begin{figure*}[ht]
\centering
  \includegraphics[width=0.7\textwidth]{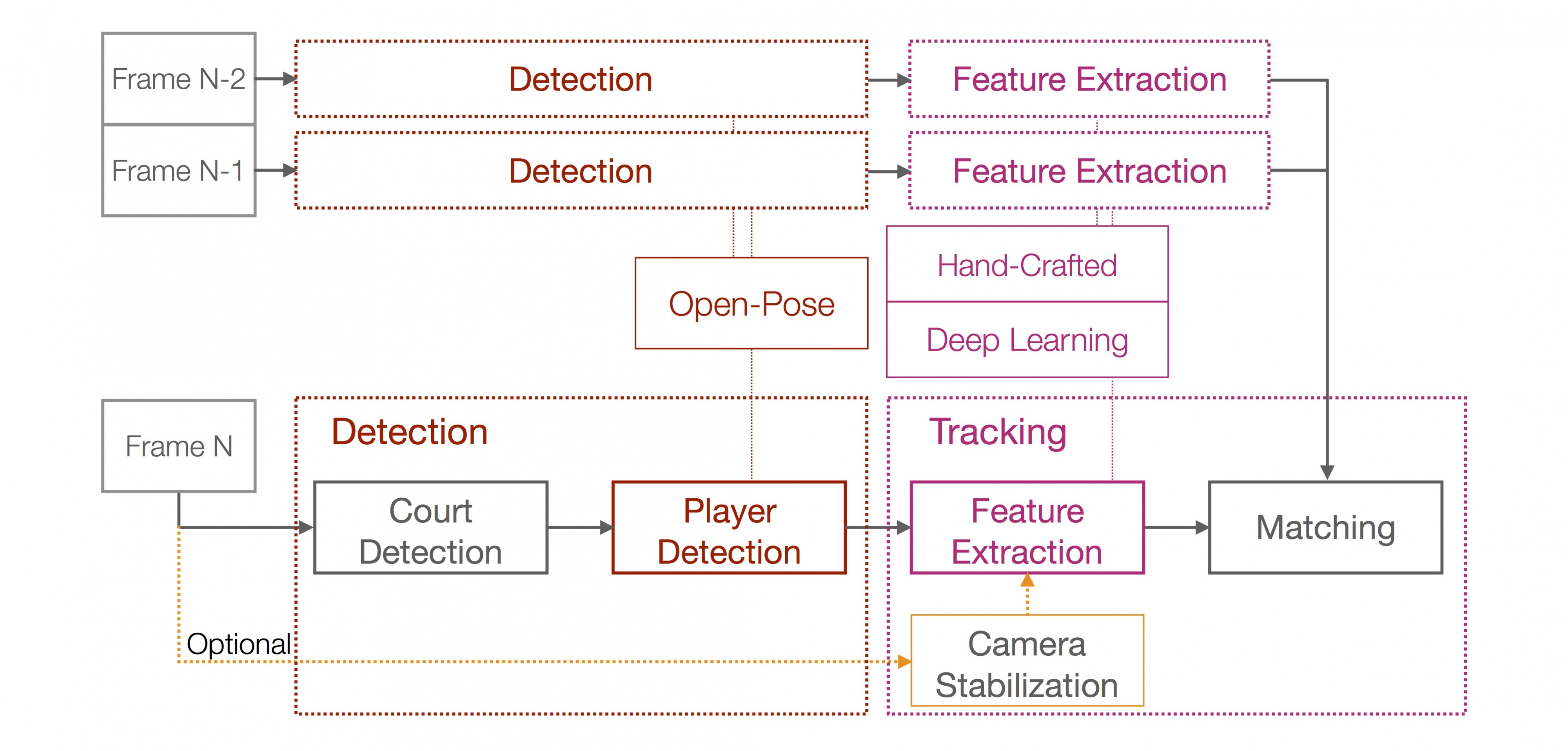}
  \caption{Generic Pipeline: for each frame, players are detected (through pose models) and tracked (via feature extraction and matching).}
  \label{fig:Pipe}
\end{figure*}

In this Section, the implemented tracking-by-detection method is detailed. The associated generic pipeline can be seen in Figure \ref{fig:Pipe} and it follows the subsequent stages: 
\begin{enumerate}
    \item[A.] For each frame, the basketball court is detected, with the purpose of not taking fans and bench players into account in the following steps. Also, a camera stabilization step may be included, and  a discussion about its need in order to perform multi-tracking by reducing distances of objects within frames is provided.
    \item[B.] Players are detected, together with their pose using a pretrained pose model, and bounding boxes are placed around all of them. 
    \item[C.] Features are extracted from these bounding boxes in combination with pose information. Several choices are analyzed in terms of features to be extracted.  
    \item[D.] By comparing features of all players in three consecutive frames (indicated by Frame N, N-1 and N-2, respectively, in Figure~\ref{fig:Pipe}) and using a customized version of the Hungarian algorithm, tracking associations are performed.
\end{enumerate}

\subsection{Pre-Processing}
\subsubsection{Court Detection}
\begin{figure}[ht]
\centering
  \includegraphics[width=0.45\textwidth]{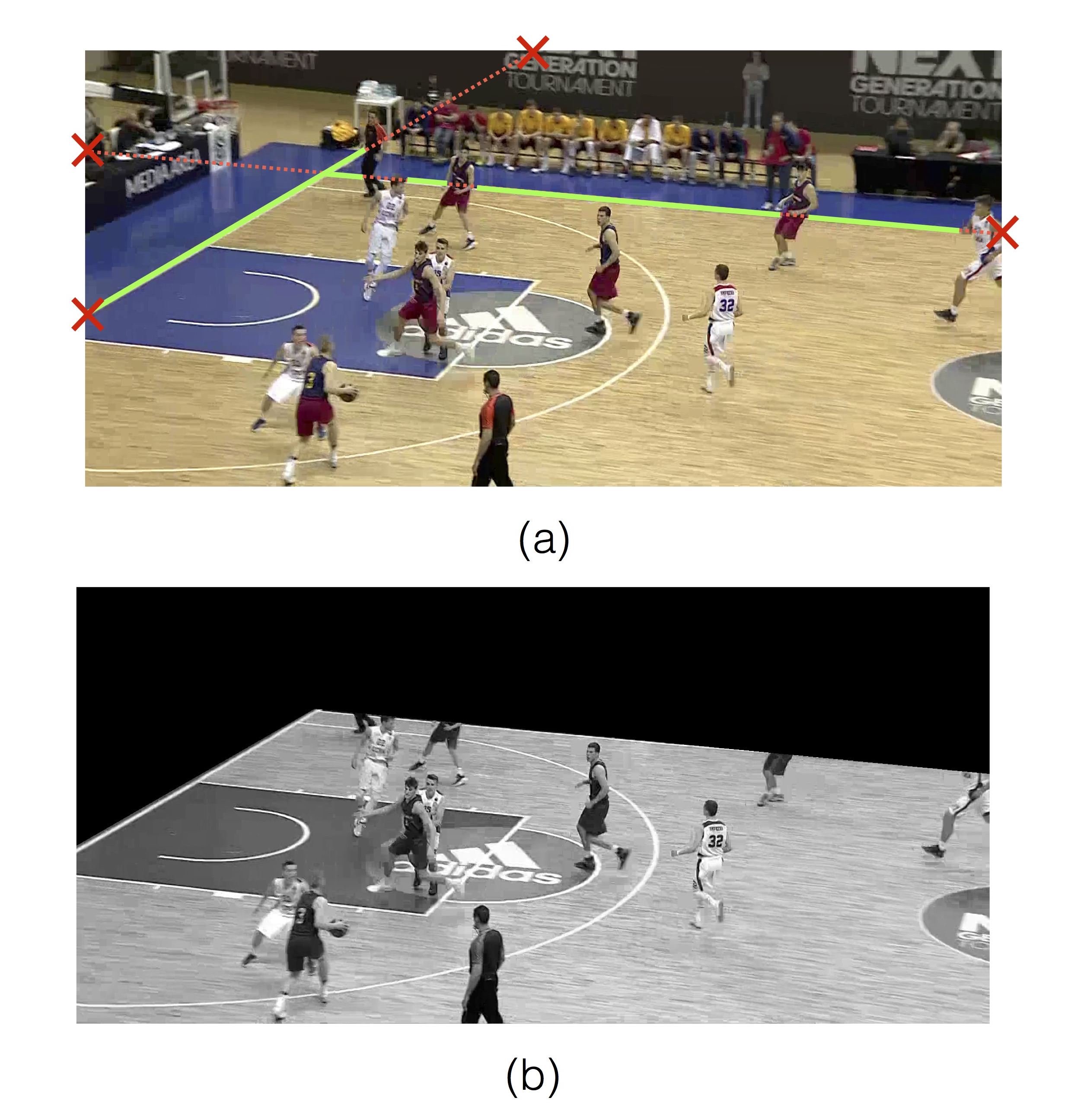}
  \caption{Court detection. (a) Different segments with the same orientation and intersections are joint; (b) Final segmentation result.}
  \label{fig:Court}
\end{figure}

Although court detection is not the main contribution of this research, the identification of visible court boundaries in the image is basic in order to filter out those candidates that are not taking part of the game actively (such as bench players or referees). It has to be mentioned that the basic filtering to be performed is thought for the vast majority of European courts, where court surroundings usually share the same color, and fans sit far from team benches. Knowing that in the broadcasting images the court results in a trapezoid with some visible boundaries, line segments are detected by using a fast parameter-less method based on the \textit{a contrario} theory \cite{von2010lsd} (code available in \cite{von2012lsd}). Right after, segments with the same orientation and intersection at the boundaries of the image, as seen in Figure \ref{fig:Court}, are joint and considered as part of the same line; the dominant orientation will be considered as the one with the longest visible parts (proportional to the sum of individual segments' length). However, given that basketball courts have many parallel lines (such as sidelines, corner three line, paint sides,...), several line candidates have to be tested in order to find the real court surroundings. Moreover, two dominant orientations are taken into account: (1) the ones belonging to sidelines (intersections at both left-right image boundaries), and (2) the one belonging to the visible baseline (both baselines cannot be seen at the same time if the camera shot is an average one).
Given the non-complex scenario of European courts, color filtering is used in the HSV colorspace by checking contributions all over the image; in the case of Figure \ref{fig:Court}, court surroundings are blue and the court itself is a bright brown tonality. For a given dominant orientation, the subsquent steps are followed: 
\begin{enumerate}
    \item First, a line candidate with the dominant orientation is set at the top (in the case of sideline candidates) /  left side (baseline candidates) of the image. 
    \item Then, two parallel lines are set at a $\pm 25$ pixel distance with respect to the line candidate.
    \item Later on, and taking only the pixels comprised between the candidate line and the two parallel ones, the number of pixels that satisfy color conditions is computed for both sides independently. This is, if the candidate line is a potential sideline, the number of pixels is computed above and under it; instead, if the candidate line is a potential baseline, the number of pixels is computed at its left and its right. In the case of Figure \ref{fig:Court}, pixels with a Hue value between 120 and 150 degrees are the ones satisfying filter conditions.
    \item The line candidate is moved 12 pixels towards the bottom (sidelines) / right side (baseline) of the image. The same procedure being followed in Steps 2 and 3 is applied again.  
    \item Once examined all possible cases, the line candidate with the maximum difference between above-below/left-right sides is set as the court limit. As it can be seen in the given example, the best court delimiters are the lines that stay right in the limit between brown-blue regions.
\end{enumerate}

\subsubsection{Camera Stabilization}
In order to ease the tracking of the players, an additional camera stabilization step to remove the camera motion can be incorporated. Taking into account that its inclusion represents extra computations, in this paper, an ablation study is provided to discuss the extend of its advantages. When enclosed, the camera stabilization method and implementation in \cite{sanchez2017ipol} is used. It estimates a set of homographies, each of which associated to a frame of the video and allowing to stabilize it. Table~\ref{tab:stab} in Section~\ref{sec:Res} presents the quantitative results including it. 

\subsection{Player Detection}

As mentioned in Section \ref{sec:Intro}, the presented tracker is based on multiple detections in each individual frame. More concretely, the implemented method relies on pose models techniques \cite{poseModel1, poseModel2, cao2017realtime} stemming from an implementation of the latter \cite{openPoseLib}. Basically, this method is a bottom-up approach that uses a Convolutional Neural Network to: (1) detect anatomical keypoints, (2) build limbs by joining keypoints, and (3) merge limbs in the visible person skeleton. Given a basketball frame, the output of the main inference pose function is a $ 25 \times 3$ vector for each player, with the position (in screen coordinates) of 25 keypoints, which belong to the main biometric human-body parts, together with a confidence score. Note that there might be situations where specific parts might not be detected, resulting in unknown information in the corresponding entry of the pose vector of the whole skeleton. In addition, 26 heatmaps are returned, indicating the confidence of each part being at each particular pixel. By checking all the parts' positions and taking the minima-maxima XY coordinates for each detected player, bounding boxes are placed around the respective players.

\subsection{Feature Extraction}
Once bounding boxes are obtained, their comparison must be performed in order to assign individual tracks for each box in time. With the purpose of quantifying this process, different approaches can be used whilst extracting features. In this subsection, all tested features used \textit{a posteriori} are explained. For the remaining part of this subsection, ${B_{t_1}}$ and ${B_{t_2}}$ are considered as two different bounding boxes, detected at $t_{1}$ and $t_{2}$ respectively.

\subsubsection{Geometrical Features} 
This classical approach can be used to measure distances or overlapping between bounding boxes of different frames. If the number of frames per second the video feed is not low, it can be assumed that player movements between adjacent frames will not be large; for this reason, players can be potentially found at a similar position in screen coordinates in short time intervals, so the distance between bounding boxes' centroids can be used as a metric. That is, given $\mathbf{x}_{B_{t_1}}$ and $\mathbf{x}_{B_{t_2}}$ as the centroids of two bounding boxes, the normalized distance between centroids can be expressed as
\begin{equation}\label{eq:Cd}
    C_d(B_{t_1},B_{t_2})=\frac{1}{\sqrt{w^2+h^2}}\|\mathbf{x}_{B_{t_1}}-\mathbf{x}_{B_2}\|,
\end{equation}
where $w$ and $h$ are  the width and the height of the frame domain. Another similar metric that could be used is the intersection over union between boxes, but due to the fact that basketball courts are usually cluttered and players move fast and randomly, it is not useful for this paper's purposes.

\subsubsection{Visual Features}\label{sec:vsim}
Distances might help distinguish basic correspondences, but this simple metric does not take into account key aspects, such as the jersey color (which team do players belong) or their skin tone. For this reason, a color similarity could be implemented in order to deal with these situations. Moreover, in this specific case, knowing that body positions are already obtained, fair comparisons can be performed, where the color surroundings of each part will be only compared to the neighborhood of the same part in another bounding box. Nevertheless, it has to be taken into account that only the pairs of anatomical keypoints present or detected in both $B_{t_1}$ and $B_{t_2}$ (denoted here as  $\mathbf{p}^k_1$ and $\mathbf{p}^k_2$, respectively) will be used for the computation. The color and texture of a keypoint can be computed by centering a neighborhood around it.  That is, let ${\cal{E}}$ be a squared neighborhood of 3$\times$3 pixels centered at $\mathbf{0}\in\mathbf{R}^2$.  Then,
\begin{equation}\label{eq:Cc} 
        C_{c}(B_{t_1},B_{t_2})\!=\!\frac{1}{255 |S| \, |\cal{E}|}\!\!\sum_{k\in S}\!\sum_{\mathbf{y}\in{\cal{E}}}\!\|I_{t_1}(\mathbf{p}^k_1+\mathbf{y})-I_{t_2}(\mathbf{p}^k_2+\mathbf{y})\|
\end{equation}
where $S$ denotes the set of mentioned pairs of corresponding keypoints detected in both frames.

\subsubsection{Deep Learning Features} \label{sec:DLF}
Deep Learning (DL) is a broadly used machine learning technique with many possible applications, such as classification, segmentation or prediction. The basis of any DL model is a deep neural network  formed by many layers. These networks serve to predict values from a given input. Convolutional Neural Networks (CNN) are special cases in which weights at every layer 
are shared spatially across an image. This has the effect of reducing the number of parameters needed for a layer and gaining a
certain robustness to translation in the image. Then, a CNN architecture is composed by several kinds of layers, being convolutional layers the most important ones, but also including non-linear activation functions, biases, etc. This type of layers computes the response of several filters by convolving with different image patches. 
The associated weights to these filters, and also the ones associated to the non-linear activation functions, are learnt during the training process (in a supervised or unsupervised way) in order to achieve maximum accuracy for the concrete aimed task. It is well known that the first convolutional layers will produce higher responses to low-level features such as edges while posterior layers correlate with mid-, high- and global-level features associated to more semantic attributes. Bearing in mind that training a model from scratch is expensive, researchers use pretrained models and their corresponding weights for their purposes, such as by fine-tuning the model (for instance by feeding the model with new data and changing or adapting the previously obtained weights accordingly). 

In the presented experiments, the popular VGG-19 network \cite{simonyan2014very} is used for feature extraction, initialized with weights trained with ImageNet dataset~\cite{imagenet_cvpr09}. The original model was trained for image classification, and its architecture consists of 5 blocks with at least 2 convolutional layers, and 2 fully-connected layers at the end that will output a class probability vector for each image. The network takes as input a $224 \times 224 \times 3$ image, and the output size of the second convolutional layer of each block is shown in Table \ref{tab:OutVGG}. 

\begin{table}[ht]
\centering
\resizebox{0.35\textwidth}{!}{
\begin{tabular}{|c|c|c|c|}
\hline
\textbf{} & \textbf{Width} & \textbf{Height} & \textbf{Nº Filters} \\ \hline
b2c2      & 112            & 112             & 128                 \\ \hline
b3c2      & 56             & 56              & 256                 \\ \hline
b4c2      & 28             & 28              & 512                 \\ \hline
b5c2      & 14             & 14              & 512                 \\ \hline
\end{tabular}}
\caption{Output size of VGG-19 convolutional layers. In the first column, b stands for block number and c stands for the convolutional layer number inside that block.}
\label{tab:OutVGG}
\end{table}

\begin{figure}[ht]
\centering
  \includegraphics[width=0.45\textwidth]{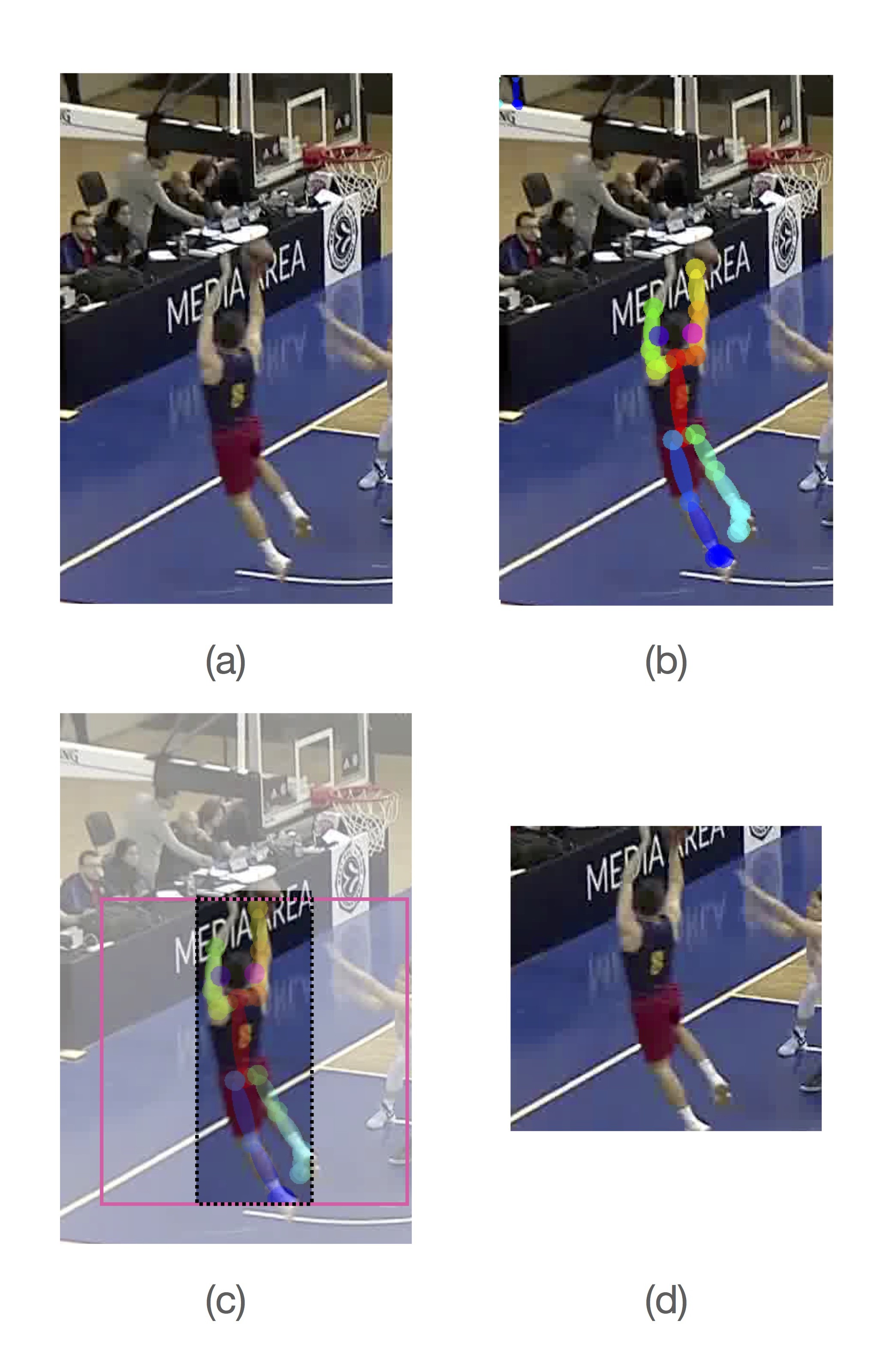}
  \caption{Player and Pose Detection: (a) random patch of an image containing a player, (b) detected pose through pretrained models, (c) black: bounding box fitting in player boundaries, pink: bounding box with default $224 \times 224$ pixels resolution, (d) reshaped bounding box to be fed into VGG-19.}
  \label{fig:resizing}
\end{figure}

In order to feed the network with an appropriate sized image, a basic procedure is followed as seen in Figure \ref{fig:resizing}: considering that player boxes are usually higher than wider and having the center of the bounding box, its height $H_{B_{t}}$ is checked. Then, a squared image of $H_{B_{t}} \times H_{B_{t}} \times 3$ is generated around the center of the bounding box; finally, this image is resized to the desired width and height ($224$ and $224$, respectively). In this way, the aspect ratio of the bounding box content does not change.  

\begin{figure*}[ht]
\centering
  \includegraphics[width=0.8\textwidth]{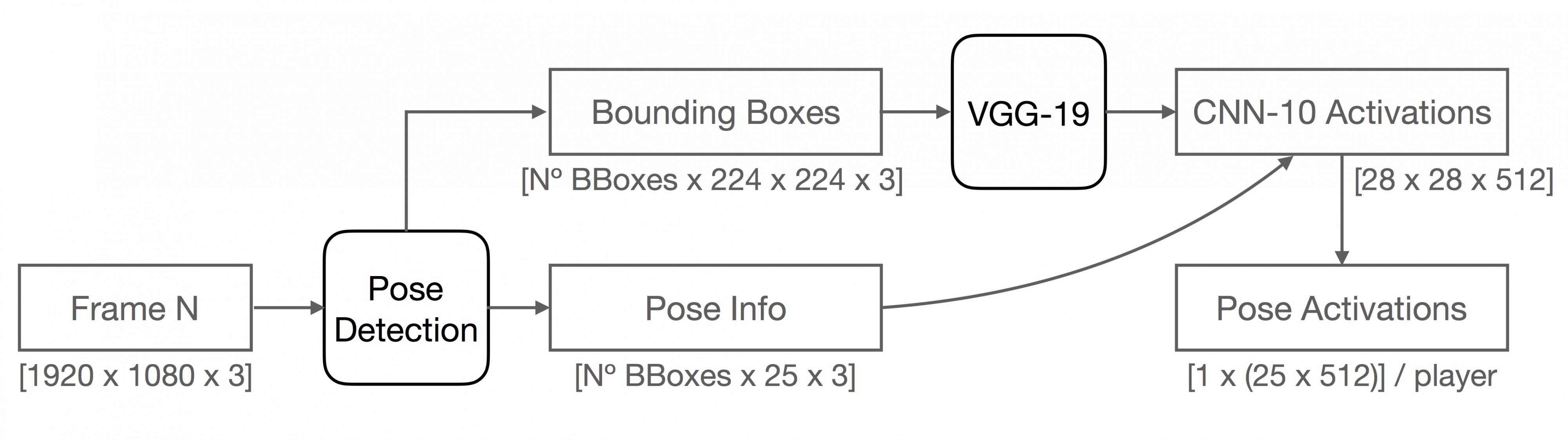}
  \caption{Feature Extraction of all body parts using the 10th convolutional layer of a VGG-19 network.}
\label{fig:featExVGG}
\end{figure*}

However, extracting deep learning features from the whole bounding box introduces noise to the feature vector, as part of it belongs to the background (e.g. court). Therefore, feature are only extracted in those pixels that belong to detected body parts, resulting in a quantized 1D vector with length equal to the number of filters. As detailed below, part positions have to be downscaled to the output size of the convolutional layer 
Moreover, all feature vectors must be normalized with L2 norm. 
An example using the 10th convolutional layer of VGG-19 is shown in Figure \ref{fig:featExVGG}, where a $1 \times (25 \times 512)$ vector is obtained.\\

Once all boxes have their corresponding feature vectors, the metric defined in \cite{wang2019learning} is used to quantify differences; in particular, the similarity between two feature vectors $f_{t_1,k}^{y_{t_1}}$ and $f_{t_2,k}^{y_{t_2}}$, belonging to bounding boxes detected in $t_{1}$ and $t_{2}$ respectively, can be defined as:
\begin{equation}\label{eq:CsSing}
    Sim(f_{t_1,k}^{y_{t_1}},f_{t_2,k}^{y_{t2}})=\frac{exp(f_{t_1,k}^{y_{t_1}} \dot f_{t_2,k}^{y_{t2}})}{\sum exp(f_{t_1,k}^{y_{t_1}} \dot f_{t_2,k}^{y_{t2}})}
\end{equation}
where $k$ corresponds to the particular body part and $y_{t1}$ and $y_{t2}$ to the pixel position inside the neighborhood being placed around the keypoint. Therefore, the total cost taking all parts into account is defined as: 
\begin{equation}\label{eq:CsTot}
     C_{DL}(B_{t_1},B_{t_2})\!=\!\frac{1}{|S|}\!\!\sum_{k \in S} \!\max_{\substack{y_{t1} \in {\cal{E}} \\ y_{t2} \in {\cal{E}}'}  }(Sim(f_{t_1,k}^{y_{t_1}},f_{t_2,k}^{y_{t2}}))
\end{equation}
where ${\cal{S}}$ corresponds, once again, to the set of detected parts in both frames, and ${\cal{E}}$ and ${\cal{E}'}$ correspond to the set of pixels in the neighborhood placed around each keypoint. 

Nevertheless, two important remarks have to be pointed out: 

\begin{enumerate}
    \item Some of the Open Pose detected parts have low confidence. Given that, generally, there are more than 14 detected parts per player, all parts with lower confidence than 0.3 are discarded and not taken into account when extracting features. Hence, the subset ${\cal{S}}$ in Equations \ref{eq:Cc} and \ref{eq:CsTot} considers all detected parts in both bounding boxes that satisfy the mentioned confidence threshold.
    \item Convolutional layer outputs 
    (as implemented in the VGG-19) decrease the spatial resolution of the input. 
    Since non-integer positions are found when downscaling parts' locations (in the input image) to the corresponding resolution of the layer of interest,  
    the features of the $2 \times 2$ closest pixels at that layer are contemplated. Then, the cost will be considered as the most similar feature vector to the $2 \times 2$ target one given. In Tables~\ref{tab:nonstab} and \ref{tab:stab} a discussion on the effect of the approximate correct location is included.
\end{enumerate}
    
\subsection{Matching}

Having quantified all bounding boxes in terms of features, a cost matrix containing the similarity between pairs of bounding boxes is computed by combining the different extraction results. The suitability of the different types of features is evaluated by combining with appropriate weights them before building this matrix; in the presented experiments, the following weighted sum of different costs has been applied: 
\begin{equation}\label{eq:cost}
C(B_{t_1},B_{t_2})= \alpha {C}_{Feat1}(B_{t_1},B_{t_2}) + (1-\alpha) {C}_{Feat2}(B_{t_1},B_{t_2})
\end{equation}
where $C_{{Feat}1}$ refers to $C_d$ given by (\ref{eq:Cd}), ${C}_{{Feat}2}$ refers either to $C_{DL}$ in (\ref{eq:CsTot}) or $C_c$ in (\ref{eq:Cc}) and $\alpha\in [0,1]$.
From this matrix, unique matchings between boxes of adjacent frames are computed by minimizing the overall cost assignment: 
\begin{enumerate}
    \item For each bounding box in time $t_{N}$, the two minimum association costs (and labels) among all the boxes in $t_{N-1}$ are stored in an  $A_{t_{N},t_{N-1}}$ matrix.
    \item If there are repeated label associations, a decision has to be made: 
    \begin{itemize}
        \item If the cost of one of the repeated associations is considerably smaller than the others (by +10\%), this same box is matched with the one in the previous frame. 
        \item If the cost of all the repeated associations is similar (less than 10\%), the box with the largest difference between its first and second minimum costs is set as the match. 
        \item In both cases, for all boxes that have not been assigned, the label of their second minimum cost is checked too. If there is no existing association with that specific label, a new match is set.
    \end{itemize}
    \item In order to provide the algorithm with some more robustness, the same procedure described in steps 1 and 2 is repeated with boxes in $t_{N}$ and $t_{N-2}$. This results in an  $A_{t_{N},t_{N-2}}$ matrix.
    \item For each single box, the minimum cost assignment for each box is checked at both $A_{t_{N},t_{N-1}}$ and $A_{t_{N},t_{N-2}}$, keeping the minimum as the final match. In this way, a 2-frame memory tolerance is introduced into the algorithm, and players that might be lost for one frame can be recovered in the following one.
    \item If there are still bounding boxes without assignations, new labels are generated, considering these as new players that appear on scene. Final labels are converted into unique identifiers, which will be later used in order to compute performance metrics.
\end{enumerate}


\section{Results} \label{sec:Res}
In this section, a detailed ablation of quantitative results is provided and discussed, comparing all the above-mentioned techniques and combinations. Besides, the content of the gathered dataset is explained. 

\subsection{Dataset}
A dataset of 22 European single-camera basketball sequences has been used. Original videos have full-HD resolution ($1920 \times 1080$ pixels) and 25 frames per second, but in order to provide fair comparisons, only 4 frames are extracted per second. The included sequences involve static offensive basketball motion, with several sets of screens/isolations; moreover, different jersey colors and skin tonalities are included. However, the court is the same European one for all situations, and there are no fast break / transition plays, as in the case where all players run from one side to the other, due to the fact that camera stabilization techniques do not handle these situations. The average duration of these sequences is 11.07 seconds, resulting in a total of 1019 frames. Ground truth data is attached in the given dataset, containing bounding boxes over each player and all 3 referees (taking the minimum visible X and Y coordinates of each individual) in every single frame (when visible); this results results in a total of 11339 boxes. 

\subsection{Quantitative Results}

\begin{figure}
\centering
  \includegraphics[width=0.45\textwidth]{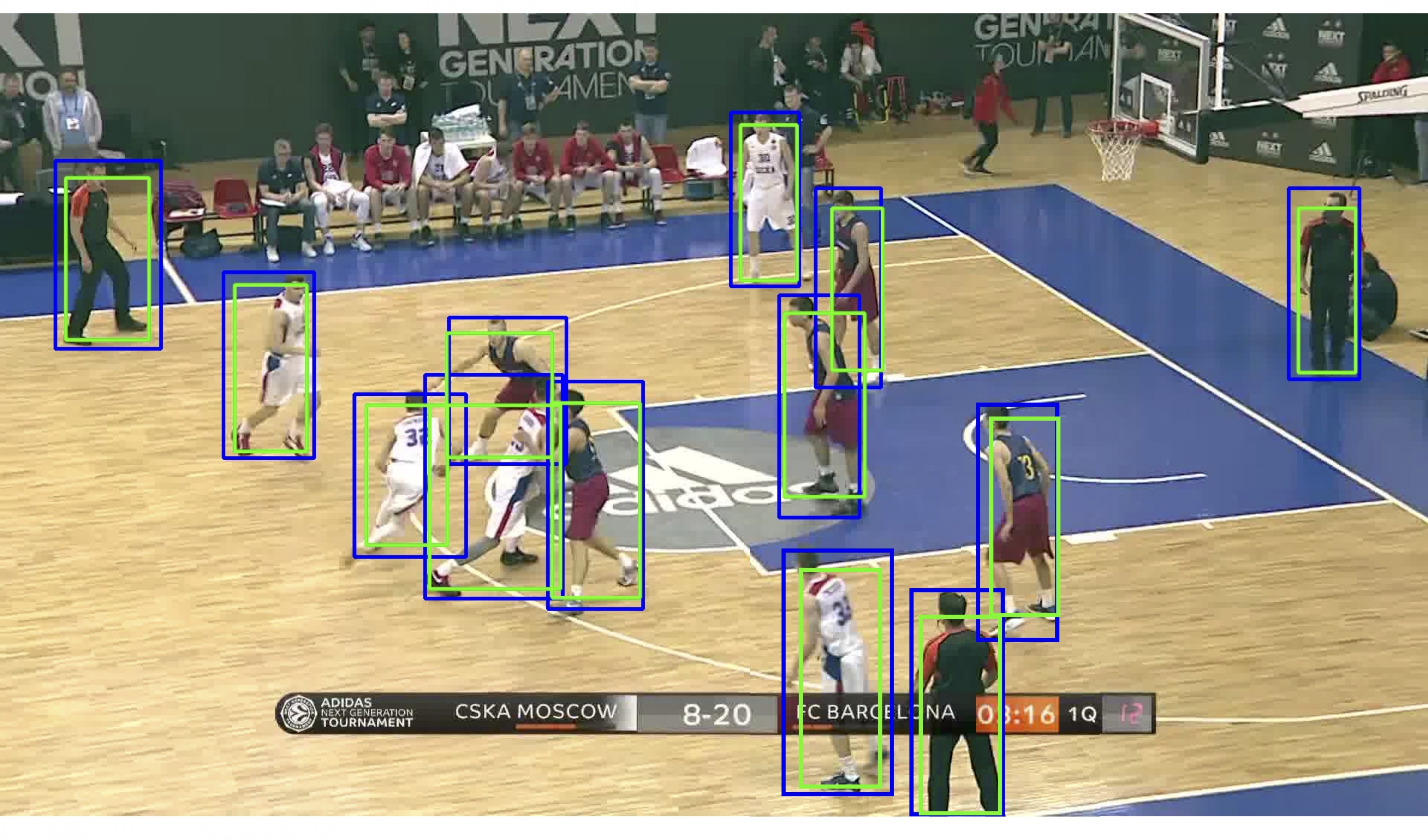}
  \caption{Player Detections (green boxes) together with its ground truth (blue boxes).}
  \label{fig:DetFig}
\end{figure}

Although it is not part of this article's contribution, quantitative assessment of the detection method is shown in Table \ref{tab:DetPlay} and it is compared to the performance of the state-of-the-art YOLO network \cite{redmon2016you}; for a fair comparison, only the \textit{person} detections within the court boundaries are kept in both cases. These detections can be seen in Figure \ref{fig:DetFig} with their corresponding ground truth boxes.  

\begin{table}[ht]
\centering
\resizebox{0.45\textwidth}{!}{
\begin{tabular}{|c|c|c|c|}
\hline
                   & \textbf{Precision} & \textbf{Recall} & \textbf{F1-Score} \\ \hline
Open Pose & 0.9718             & 0.9243          & 0.9470            \\ \hline
YOLO      & 0.8401             & 0.9426          & 0.8876            \\ \hline
\end{tabular}}
\label{tab:DetPlay}
\caption{Detection Results}
\end{table}

From now on, all quantitative tracking results will be expressed in the Multiple Object Tracking Accuracy (MOTA) metric, which is defined in \cite{bernardin2008evaluating} as: 
$$ MOTA = 1-\frac{\sum_{t}^{} fp_{t} + m_{t} + mm_{t}}{\sum_{t}^{} g_{t}},$$ 
 where $fp_{t}$, $m_{t}$, $mm_{t}$ and $g_{t}$ denote, respectively, to false positives, misses, missmatches and total number of ground truth boxes over all the sequence. 
 
Another meaningful tracking metric that has been computed as well is the Multiple Object Tracking Precision (MOTP), which can be defined as: 
$$ MOTP = \frac{\sum_{i,t}^{} IoU_{i,t}}{\sum_{t}^{} c_{t}},$$ 
 where $IoU_{i,t}$ and $\sum_{t}^{} c_{t}$ correspond to the intersection over union between two boxes, and to the sum of correct assignations through the sequence, respectively. The detected bounding boxes for all the upcoming experiments are the same ones (thus the intersection with Ground Truth bounding boxes does not change neither), and knowing that the total number of instances is large, the MOTP results barely changes in all presented combinations of techniques: $0.6165 \pm 0.0218 $. \\

Starting only with DL features (that is, $\alpha=0$ in (\ref{eq:cost}) and ${C}_{{Feat}2}$ equal to $C_{DL}$), Table \ref{table:OutConv} shows the maximum MOTA metrics achieved after performing the extraction in the output of convolutional layers. As mentioned, a pretrained VGG-19 architecture is used, taking as an output the result of each second convolutional layer from the second to the fifth block. The best MOTA results are obtained with the output of the fourth block, corresponding to the 10th convolutional layer of the overall architecture. For the remaining tests, all DL features will be based on this layer, which has an output of size $ 28 \times 28 \times 512 $. 

\begin{table}[ht]
\centering
\resizebox{0.45\textwidth}{!}{
\begin{tabular}{|c|c|c|c|c|}
\hline
\textbf{Layer} & b2c2   & b3c2   & b4c2   & b5c2   \\ \hline
\textbf{MOTA}  & 0.5396 & 0.5972 & \textbf{0.6369} & 0.6321 \\ \hline
\end{tabular}}
\caption{MOTA results obtained with $\alpha=0$ in (\ref{eq:cost}), ${C}_{{Feat}2}$ equal to $C_{DL}$ and by extracting features in the output of different convolutional layers.}\label{table:OutConv}
\end{table}

Having tried all possible weights in 0.05 intervals, Table \ref{tab:nonstab} shows the most significant MOTA results for a non-stabilized video sequence. In this experiment, a comparison between Geometrical and DL features is shown, with the performance on their own as well as its best weighted combination. Besides, as explained in Subsection \ref{sec:DLF}, when extracting DL features, three different tests have been performed regarding the neighborhood size. As it can be seen in Table \ref{tab:nonstab}, DL features outperform Geometrical ones, specially in the case of a 2x2 neighborhood. By combining them, and giving more weight to the DL side, results are improved in all cases, thus indicating that the two types of features complement each other. In Table \ref{tab:stab} the same experiments are shown, but this time using a stabilizied video sequence. In this case, the Geometrical performance outperforms Deep Learning, but as it has been mentioned, these metrics will drastically drop if the included dataset sequences contain fast camera movements (or even large pannings). 

From both Tables \ref{tab:nonstab} and \ref{tab:stab} it can be deduced that the best filter size when extracting DL pose features is a 2x2 neighborhood. \textit{A priori}, one might think that a 3x3 neighborhood should work better, as it is already including the 2x2 one, but a 3x3 spatial neighborhood in the output of the 10th convolutional layer is equivalent to a $24 \times 24$ real neighborhood around the specific part in the image domain. Accordingly, adding these extra positions will include court pixels in all feature vectors, which might then produce a higher response in court-court comparisons, resulting in non-meaningful matches. 

\begin{table}[ht]
\centering
\resizebox{0.35\textwidth}{!}{
\begin{tabular}{|c|c|c|c|}
\hline
\textbf{Neighborhood} & ${\alpha}$ &  1-$\alpha$ & \textbf{MOTA}   \\ \hline
---      & 1                    & 0                   & 0.5689          \\ \hline
1x1                   & 0                    & 1                   & 0.5923          \\ \hline
1x1                   & 0.3                  & 0.7                 & 0.6289          \\ \hline
2x2                   & 0                    & 1                   & 0.6369          \\ \hline
2x2                   & 0.2                  & 0.8                 & \textbf{0.6529} \\ \hline
3x3                   & 0                    & 1                   & 0.6171          \\ \hline
3x3                   & 0.3                  & 0.7                 & 0.6444          \\ \hline
\end{tabular}}
\caption{Non-stabilized results obtained from only 4 video frames per second.}
\label{tab:nonstab}
\end{table}

\begin{table}[ht]
\centering
\resizebox{0.35\textwidth}{!}{
\begin{tabular}{|c|c|c|c|}
\hline
\textbf{Neighborhood} & $\alpha$ &  1-$\alpha$ & \textbf{MOTA}    \\ \hline
---        & 1                    & 0                   & 0.6506          \\ \hline
2x2                   & 0                    & 1                   & 0.6369          \\ \hline
1x1                   & 0.6                  & 0.4                 & 0.6752          \\ \hline
2x2                   & 0.55                 & 0.45                & \textbf{0.6825} \\ \hline
3x3                   & 0.7                  & 0.3                 & 0.6781          \\ \hline
\end{tabular}}
\caption{Stabilized results, with the same 4 video frames per second and weights as in Table~\ref{tab:nonstab}.}\label{tab:stab}
\end{table}

Apart from comparing Geometrical and DL features through $C_d$ and the mentioned different $C_{DL}$, the effect of Visual features (color similarity $C_c$, explained in Subsection \ref{sec:vsim}) is checked too. In Table \ref{tab:ConDLFeat}, the best weighted combinations in terms of MOTA are shown for a non-stabilized and a stabilized video sequence. In both cases, DL features outperform color ones by a 3\% margin. The combination of all Geometrical, Visual, and DL features outperforms the rest of techniques but just by a 0.2\%, which comes at a cost of computation expenses, so it is worth using only DL features. \\

\begin{table}[ht]
\centering
\resizebox{0.4\textwidth}{!}{
\begin{tabular}{|c|c|}
\hline
\textbf{Combination of Features}    & \textbf{MOTA}   \\ \hline
Geometrical + Visual            & 0.6233          \\ \hline
Geometrical + VGG                   & 0.6529          \\ \hline
Geometrical + Visual {[}Stab{]} & 0.6583          \\ \hline
Geometrical + VGG {[}Stab{]}        & 0.6825 \\ \hline
Geometrical + VGG + Visual {[}Stab{]}        & \textbf{0.6843} \\ \hline
\end{tabular}}
\caption{Effect of Visual and Deep Learning features in combination with Geometrical ones.}
\label{tab:ConDLFeat}
\end{table}

In order to break down and evaluate the contribution in MOTA of every single pose part, Table \ref{tab:partPerf} is displayed; these results have been obtained with a 2x2 neighborhood around parts, and without combining with Geometrical features. As it can be seen, there are basically three clusters: 
\begin{enumerate}
\item Discriminative features, above a 0.35 MOTA, that manage to track at a decent performance only with a $1 \times 512$ feature vector/player. These parts (shoulders, chest and hip) belong to the main shape of human \textit{torso}, and it coincides with the jersey-skin boundary in the case of players.
\item Features that stay within a MOTA of 0.20 and 0.35, which are not tracking players properly but their contribution might help the discriminative ones to get higher performance metrics. These parts include skinned pixels of basic articulations such as elbows, knees, and ankles. 
\item Concrete parts that have almost no details at a coarse resolution, thus resulting in low MOTA performance. Eyes could be an example: although people's eyes have many features that made them discriminative (such as shape, color, pupil size, eyebrow's length), players' eyes in the dataset images do not embrace more than a 5x5 pixel region, and all of them look the same shape and brown or darkish. This results in poor tracking results when checking only for these parts.
\end{enumerate}

\begin{table}[ht]
\centering
\begin{tabular}{|c|c|}
\hline
\textbf{Part} & \textbf{MOTA} \\ \hline
Chest         & 0.5349        \\ \hline
L-Shoulder    & 0.4726        \\ \hline
R-Shoulder    & 0.4707        \\ \hline
R-Hip         & 0.3961        \\ \hline
Mid-Hip       & 0.3956        \\ \hline
L-Hip         & 0.3867        \\ \hline
L-Knee        & 0.3156        \\ \hline
R-Knee        & 0.3062        \\ \hline
L-Elbow       & 0.2862        \\ \hline
R-Elbow       & 0.2545        \\ \hline
R-Ankle       & 0.2418        \\ \hline
L-Ankle       & 0.2407        \\ \hline
L-Toes        & 0.1935        \\ \hline
R-Toes        & 0.1920        \\ \hline
L-Ear         & 0.1348        \\ \hline
L-Heel        & 0.1259        \\ \hline
L-Wrist       & 0.1235        \\ \hline
R-Heel        & 0.1126        \\ \hline
L-Mid-Foot    & 0.1116        \\ \hline
R-Wrist       & 0.1111        \\ \hline
R-Mid-Foot    & 0.0964        \\ \hline
L-Eye         & 0.0916        \\ \hline
Nose          & 0.0771        \\ \hline
R-Eye         & 0.0655        \\ \hline
R-Ear         & 0.0677        \\ \hline
\end{tabular}
\caption{Individual Part Tracking Performance, obtained with $\alpha=0$ in (\ref{eq:cost}) and ${C}_{{Feat}2}$ equal to $C_{DL}$}
\label{tab:partPerf}
\end{table}

Given the mentioned clusters, 3 different tracking tests have been performed taking only some parts into account, in particular, taking all the body parts that had a MOTA performance by itself higher than: (1) 0.35, (2) 0.20, (3) 0.10, belonging to (1) 6, (2) 12 and (3) 20 parts, respectively. Results are shown in Table \ref{tab:sepParts}, where it can be seen that the second and third cluster complement the top ones, while the bottom-5 parts actually contribute to a drop in MOTA. The drawback of this clustering is that it requires some analysis that cannot be performed in test time, and different video sequences (\textit{i.e} different sports) might lead to different part results. 

\begin{table}[ht]
\centering
\resizebox{0.4\textwidth}{!}{
\begin{tabular}{|c|c|c|}
\hline
\textbf{Part MOTA} & \textbf{Nº of Parts} & \textbf{Total MOTA} \\ \hline
\textgreater 0.35  & 6                    & 0.6105              \\ \hline
\textgreater 0.20  & 12                   & 0.6412              \\ \hline
\textgreater 0.10   & 20                   & \textbf{0.6423}     \\ \hline
all                  & 25                   & 0.6369              \\ \hline
\end{tabular}}
\caption{Clustering Part Results ($\alpha=0$ and ${C}_{{Feat}2}=C_{DL}$)}
\label{tab:sepParts}
\end{table}

\begin{figure*}[ht]
\centering
  \includegraphics[width=0.95\textwidth]{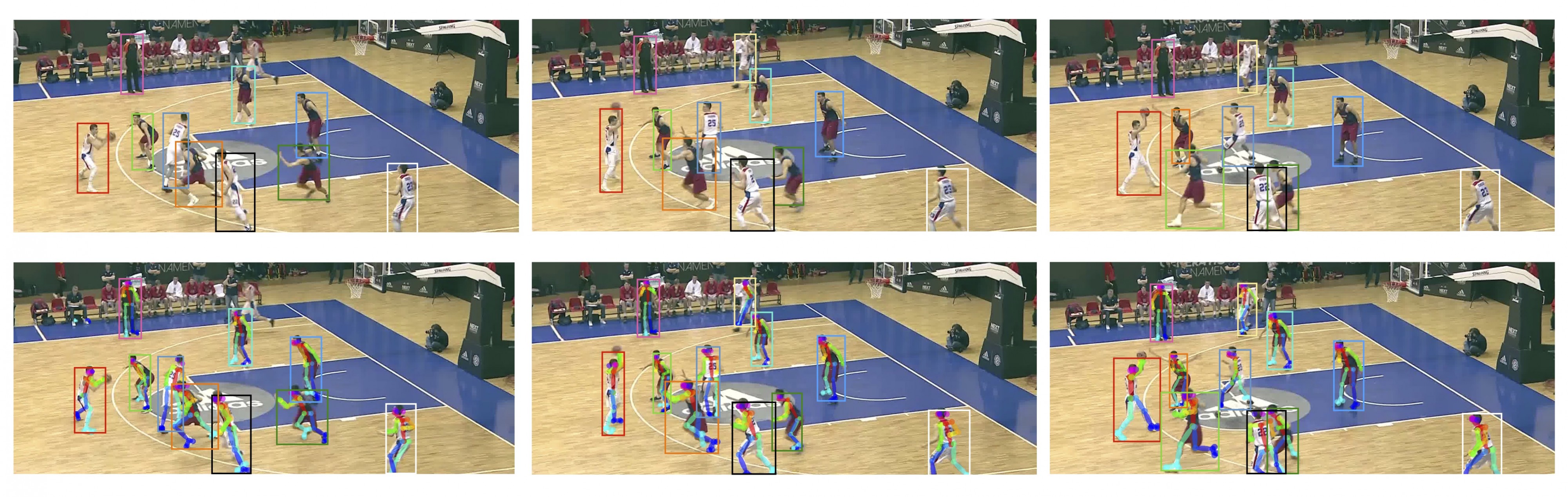}
  \caption{Obtained tracking and pose results in three consecutive frames, where each bounding box color represents a unique ID.}
  \label{fig:FinRes}
\end{figure*}

A qualitative visual detection and tracking result (obtained with the best combination of Geometrical + Deep Learning features without camera stabilization) is displayed in Figure \ref{fig:FinRes}, where players are detected inside a bounding box, and its color indicates their ID; as it can be seen, all 33 associations are properly done except a missed player in the first frame and a missmatch between frames 2 and 3 (orange-green boxes)  

\section{Conclusions} \label{sec:Conc}
In this article, a single-camera multi-tracker for basketball video sequences has been presented. Using a pretrained model to detect pose and humans, an ablation study has been detailed in order to address the feature extraction process, considering three types of features: Geometrical, Visual and Deep Learning based. In particular, Deep Learning features have been extracted by combining pose information with the output of convolutional layers of a VGG-19 network, reaching a maximum of 0.6843 MOTA performance. Several conclusions can be extracted from the presented experiments: 
\begin{itemize}
    \item In the case of VGG-19, DL features extracted from the 10th convolutional layer present the best accuracy; moreover, placing a 2x2 neighborhood around downscaled body parts improves the tracking performance.  
    \item Classical Computer Vision techniques such as camera stabilization can help improving the Geometrical features performance, but it might have related drawbacks, such as the incapability of generalization to all kinds of camera movements.
    \item DL features outperfom Visual ones when combining with Geometrical information. The combination of all of them does not imply a performance boost. 
    \item When extracting pose features from convolutional layers, those body parts that cannot be distinguishable at a coarse resolution have a negative effect in the overall performance. 
\end{itemize}
Future work could involve the fine-tuning of a given network in order to get specific weights for tracking purposes. This training should be done in an unsupervised/self-supervised way, and a bigger dataset  will be used, including different type of basketball courts and all kind of plays. Moreover, if there is no need to label ground truth data, this new model could be also trained with other sports' data, thus potentially creating a robust multi-sport tracker. 

\section*{Acknowledgments}
The authors acknowledge partial support by MICINN/FEDER UE project, reference PGC2018-098625-B-I00, H2020-MSCA-RISE-2017 project, reference 777826 NoMADS and F.C. Barcelona's data support.

\ifCLASSOPTIONcaptionsoff
  \newpage
\fi

\end{document}